\documentclass[letterpaper, 10 pt, conference]{ieeeconf}  

\IEEEoverridecommandlockouts                              

\usepackage{epsfig} 
\usepackage{times} 
\usepackage{amsmath} 
\usepackage{amssymb}  
\usepackage{booktabs}
\usepackage{bm}
\usepackage{xcolor}
\usepackage{bbold}
\usepackage{hyperref}
\usepackage{pifont}
\usepackage{array}
\usepackage{amsfonts}
\usepackage[normalem]{ulem}

\usepackage{siunitx}
\usepackage{graphicx}
\usepackage[caption=false,font=footnotesize]{subfig}
\usepackage{cleveref}
\usepackage{tabularx}
\usepackage{makecell}
\usepackage{wrapfig}

\newcommand{\R}{\mathbb{R}}
\newcommand{\norm}[1]{\left\lVert#1\right\rVert}
\newcolumntype{P}[1]{>{\centering\arraybackslash}p{#1}}
\newcommand{\mO}{\mathcal{O}}

\newcommand{\eg}{\emph{e.g.,} } 

\def\1{\mathbb{1}}

\title{\LARGE \bf
Collision Avoidance and Navigation for a Quadrotor Swarm Using End-to-end Deep Reinforcement Learning
}

\author{Zhehui Huang*, Zhaojing Yang*, Rahul Krupani, Bask{\i}n \c{S}enba\c{s}lar, Sumeet Batra, Gaurav S. Sukhatme  
\thanks{* Equal contribution. The authors are with the Department of Computer Science, University of Southern California, Los Angeles, CA 90089 USA (e-mail: zhehuihu@usc.edu). 
GSS holds concurrent appointments as a Professor at USC and as an Amazon Scholar. This paper describes work performed at USC and is not associated with Amazon. 
}
}

\begin{document}

\maketitle
\thispagestyle{empty}
\pagestyle{empty}

\begin{abstract}
End-to-end deep reinforcement learning (DRL) for quadrotor control promises many benefits -- easy deployment, task generalization and real-time execution capability. Prior end-to-end DRL-based methods have showcased the ability to deploy learned controllers onto single quadrotors or quadrotor teams maneuvering in simple, obstacle-free environments. 
However, the addition of obstacles increases the number of possible interactions exponentially, thereby increasing the difficulty of training RL policies.
In this work, we propose an end-to-end DRL approach to control quadrotor swarms in environments with obstacles.
We provide our agents a curriculum and a replay buffer of the clipped collision episodes to improve performance in obstacle-rich environments.
We implement an attention mechanism to attend to the neighbor robots and obstacle interactions - the first successful demonstration of this mechanism on policies for swarm behavior deployed on severely compute-constrained hardware.
Our work is the first work that demonstrates the possibility of learning neighbor-avoiding and obstacle-avoiding control policies trained with end-to-end DRL that transfers zero-shot to real quadrotors. 
Our approach scales to 32 robots with $80 \%$ obstacle density in simulation and 8 robots with $20 \%$ obstacle density in physical deployment. 
\textbf{Website: \url{https://sites.google.com/view/obst-avoid-swarm-rl}}\looseness=-1

\end{abstract}

\section{Introduction}

Collision avoidance in point-to-point navigation of quadrotors is an enabler for many applications, including package delivery~\cite{dandrea2014delivery}, surveillance~\cite{borkar2020reconfigurable}, warehouse stocktaking~\cite{liu2021stock}, and search and rescue~\cite{almurib2011sr}. 
Existing real-time trajectory planning approaches~\cite{zhou2017bvc,senbaslar2019dars,senbaslar2023rlss,senbaslar2022async,tordesillas2020mader,kondo2022robust,luis2020dmpc,wang2021dpmc, park2021rsfc} are generally compute-heavy, which limits their collision avoidance ability on embedded hardware given their low reactivity. 
Existing classical collision-avoiding control methods~\cite{wang2017safety, alonsomora2013orca}, while less compute-heavy compared with real-time trajectory planning approaches, are generally conservative, which limits their performance and scalability in complex environments.\looseness=-1

In our prior work~\cite{batra2022decentralized}, we explored using end-to-end RL to train quadrotor teams to learn emergent cooperative behaviors and agile maneuvers in \textit{obstacle-free environments}.
However, the addition of obstacles poses a significant challenge, as the number of agent-environment interactions increases exponentially, thus destabilizing training in the early critical stages of learning.
In this work, we propose several changes that not only enable learning of the same agile control policies and emergent cooperative behaviors in obstacle dense environments, but also enable learning to fly through narrow gaps and generalizes to unseen scenarios. The contributions of our work are as follows:
\begin{itemize}
\item To the best of our knowledge, our approach is the first \emph{purely end-to-end DRL-based} approach that generates decentralized, low-level control policies for quadrotor swarms in obstacle-rich environments. 
The learned control policies allows reaching to goal positions while avoiding collisions with other quadrotors and static obstacles with a high success rate.
The robots are able to fly through as small as $\SI{0.15}{m}$ gaps between obstacles where the radius of the quadrotors is $\SI{0.05}{m}$. 
The learned policies are zero-shot transferrable to physical quadrotor swarms.
We utilize signed distance field (SDF) based obstacle observations, which are quantity and permutation invariant, and show their effectiveness in learning collision avoidance.
We propose a simple but useful replay mechanism, which shows its effectiveness in training and better than prioritized level replay (PLR)~\cite{jiang2021prioritized}.\looseness=-1

\item We compare our approach with state-of-the-art learning based and classical control based collision avoidance methods, and show its superior performance to the learning based and comparable performance to the classical controller using less computation time.

\item We deploy our approach to compute-constrained hardware, i.e., Crazyflie 2.1, to show its applicability to real world robots. Our work is the first to provide a successful demonstration of deploying the attention mechanism on such compute-constrained hardware.
\end{itemize}

\begin{figure*}
  \centering
    \vspace{0.08in}
    {\includegraphics[width=0.98\textwidth]{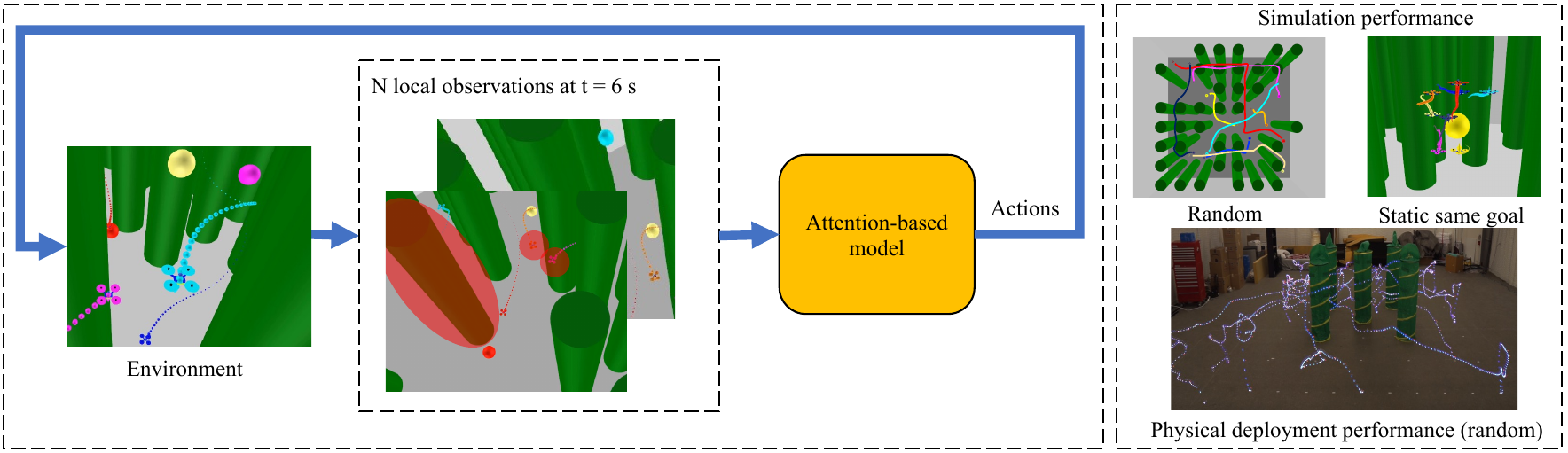}}
    \vspace{-0.12in}
    \caption{\textbf{System overview.} There are $N$ robots in the environment, and the green cylinders represent obstacles. At every tick, each robot collects its own local observations from the environment and computes its actions independently $\eg$ red shadows in the stacked local observations denote the local observations of the red robot. Our learned policy is effective in simulated trials, scales, and can be transferred to physical, severely compute-constrained quadrotors.\looseness=-1}
  \label{fig:sys_overview}
  \vspace{-0.28in}
\end{figure*}

\section{Related Work}
\label{sec:releated_work}
Compared with navigating multiple robots in obstacle-free environments, navigating multiple robots in static obstacle-rich environments significantly increases the complexity of the problem for both classical and learned methods~\cite{zhou2022swarm}. 
Its discrete variant studied as the multi-agent path finding problem is NP-Hard for total arrival time, makespan, or distance optimization~\cite{yu2013structure}. 
Its continuous variants for only geometric planning are known to be PSPACE-Hard~\cite{hopcroft1984complexity}. 
We are not only interested in kinematics but also the dynamics of the underlying systems, making the problem considerably harder. \looseness=-1

Decentralized real-time trajectory planning is utilized for multi-robot navigation in which robots plan trajectories for themselves in a receding horizon fashion avoiding collisions with each other and obstacles in the environment.
Some decentralized real-time trajectory planning algorithms require position only sensing~\cite{zhou2017bvc, senbaslar2019dars,senbaslar2023rlss} while others rely on full state sensing~\cite{wang2021dpmc, park2021rsfc} and full feature plan communication~\cite{tordesillas2020mader, luis2020dmpc}. 
Some explicitly account for asynchronous planning between robots~\cite{senbaslar2022async,kondo2022robust}.
For dynamic feasibility, some approaches formulate the problem in a model-predictive control style~\cite{zhou2017bvc,senbaslar2022async,luis2020dmpc,wang2021dpmc}, while others exploit the differential flatness of the robots~\cite{senbaslar2019dars,senbaslar2023rlss,tordesillas2020mader,kondo2022robust,park2021rsfc}.
Since these approaches make relatively longer horizon decisions, they typically require considerable computational resources, making them inadequate for computationally limited hardware. 
For example, each planning iteration in MADER~\cite{tordesillas2020mader} takes $\sim200ms$ on Intel i7 processors @4.7 GHz with access to a state-of-the-art optimization library, Gurobi, using a considerable amount of memory, and RSFC~\cite{park2021rsfc} takes $\sim50ms$ per iteration on Intel i7 processors @ 3.60 GHz using CPLEX optimization studio, both of which are state-of-the-art online planning methods.\looseness=-1

To execute on computationally limited hardware, some decentralized approaches compute only the next actions to execute at high frequency, such that if all robots compute their next action using the same approach, the system stays collision-free.
ORCA~\cite{alonsomora2013orca} computes safe velocity commands, SBC~\cite{wang2017safety}, which is based on safety barrier certificates, computes safe acceleration commands, both of which utilize real-time optimization to compute the next actions. However, these algorithms are generally more conservative than the planners, which limits their performance and scalability in complex environments.\looseness=-1

Utilizing learning while computing next actions has been investigated in order to tackle conservativeness short horizon approaches.
GLAS~\cite{riviere2020glas} utilizes imitation learning to mimic a centralized planner and combines it with a safety module to make the computed actions safe. 
In~\cite{feng2021collision}, DRL is utilized for mapping observations to linear and angular velocity commands for omnidirectional robots.
In~\cite{chen2017drl}, local neighbor observations are mapped to velocity commands using DRL. 
In~\cite{cui2022safe}, a control network is learned with a control barrier function-based safety module, in which the behavior of the integrated system is considered during training.~\cite{hua2022learn} utilizes a trained policy that generates human-like commands (faster/slower), which are used by the downstream traditional trajectory optimizer, 
and~\cite{yan2023nn} uses DRL to control flocking fixed-wing UAVs in a leader-follower setup in which control actions are velocities and roll angles.
Since algorithms compute only the next actions to execute, they can run in limited capability hardware.\looseness=-1

Using a learning-based method to train neural network policies that directly map local observations to rotor thrusts can correctly represents the true actuation limits of the quadrotor platform given there is no additional control loop needed, and potentially execute much more agile and less conservative maneuvers compared with high-level control inputs, such as desired linear velocity or future waypoints~\cite{kaufmann2022benchmark}.
\cite{batra2022decentralized} proposes an end-to-end DRL-based approach to control quadrotors with thrust inputs in obstacle-free environments. 
The proposed approach, trained in simulation~\cite{huang2023quadswarm}, can be zero-shot transferred to the real robots. However, directly applying this approach to obstacle-rich environments does not result in acceptable performance as we show in~\autoref{sec:experiments_and_results}. We address the limitations in~\cite{batra2022decentralized} and show collision avoidance and team navigation in obstacle-rich environments. Our approach, based on~\cite{batra2022decentralized, molchanov2019sim}, inherits the features of zero-shot transferable, robust to external disturbances, can withstand harsh initial conditions, and recoverable from collisions.\looseness=-1

\section{Method}
\label{sec:method}

\subsection{Problem Formulation}
The state of the environment with $N$ robots and $M$ static obstacles at time $t$ is $(s_{1}^{t}, 
\ldots, s_{N}^{t}, g_{1}^{t}, \ldots, g_{N}^{t}, 
\mO_{1}^{t}, 
\ldots, \mO_{M}^{t})$, where 
$s_{i}^{t}$ is the state of \mbox{robot $i$}, containing the position, linear and angular velocity, and orientation of the robot, 
$g_{i}^{t}$ is the goal position of robot $i$, and 
$\mO_{i}^{t}$ is the state of the static obstacle $i$, containing the position of the obstacle at time $t$.
The actions are rotor thrusts, which affect the states of the robots according to quadrotor dynamics~\cite{mellinger2011snap,lee2010control}.
Our objective is to train a decentralized control policy that directly maps local observations of a robot, which mentioned in~\autoref{subsec:training_setup}, to its rotor thrusts with the goal of minimizing its distance to its goal while avoiding collisions with other robots and static obstacles.\looseness=-1

\subsection{Training Setup}
\label{subsec:training_setup}
We train our control policies in a $10m\times10m\times10m$ simulated room with obstacles, where the height of the obstacles is the same as the room height. 
We discretize the center $8m\times8m$ area of the room to $64$ square grid cells of $1m^2$, and spawn obstacles at the centers of the cells.
At the beginning of each training episode, we randomly generate obstacles with a configurable density of occupied grid cells and obstacle sizes.
Following this, we spawn robots in the centers of obstacle-free grid cells at random heights between $1m$ and $3m$ with random initial orientations and velocities.\looseness=-1

\textbf{Goal generation}: 
We use two goal generation methods.
In the first, all robots share the same static goal position. 
The goal is spawned at the position within the room that is farthest away from any obstacles. 
The robots need to navigate around obstacles as they move toward the goal, and they need to avoid each other while staying close to the goal.
In the second, each robot has an uncorrelated randomly generated goal. 
Robots need to navigate to the goals, while avoiding collisions with obstacles and other robots.\looseness=-1

\textbf{Observations and actions}: 
The observation of robot $i$ at time t is: 
$o_{i}^{t} = (e_{i}^{t}, \eta_{i}^{t}, \zeta_{i}^{t})$, where 
i) $e_{i}^{t}$ is robot's observation of its own state and goal, 
ii) $\eta_{i}^{t}$ is observation of the neighbor robots, and 
iii) $\zeta_{i}^{t}$ is the observations of obstacles.
Specifically,
$e_{i}^{t} = (p_i^t, v_i^t, R_i^t, \omega_i^t, h_i^t)$, where 
$p_i^t \in \R^3$ is the position of the robot relative to its goal position, 
$v_i^t \in \R^3$ is its linear velocity in the world frame,  
$R_i^t \in SO(3)$ is the rotation matrix from the body frame to the world frame, $\omega_i^t \in \R^3$ is its angular velocity in the body frame, and $h_i^t \in \R$ is the altitude of robot in the room.
$\eta_{i}^{t} = (\Tilde{p}_{i1}^{t}, \ldots, \Tilde{p}_{iK}^{t}, \Tilde{v}_{i1}^{t}, \ldots, \Tilde{v}_{iK}^{t})$, where
$\Tilde{p}_{ij}^{t} \in \R^{3}$ and $\Tilde{v}_{ij}^{t} \in \R^{3}$ are the position and velocity of the robot relative to its $j$-th nearest neighbor robot, and $K \leq N - 1$ is the number of neighbors that the robot can sense. 
The obstacle observations are based on the idea of a SDF~\cite{felzenszwalb2012distance}.
The obstacle observations $\zeta_{i}^{t}$ have 9 values, which represent the distance to the closest obstacles scaled to a 3$\times$3 cells and discretized into uniformly spaced cells with the pre-defined resolution, 0.1m in our setting. The obstacle observations are quantity and permutation invariant, which can support an arbitrary number of obstacles. 
The action of robot $i$ at time t is $a_{i}^{t} \in [0,1]^4$, corresponding to the thrust levels at each of the four rotors. We transform $a_{i}^{t}$ to thrusts $f_i^t$ linearly such that $0$ is no thrust, and $1$ is maximum thrust.\looseness=-1

\textbf{Reward function}:
We extend the reward function proposed in~\cite{batra2022decentralized} for inter-robot collision avoidance in order to provide obstacle avoidance behavior as well.
The reward function for each robot $i$ consists three main components: 
$r_i^{t} = r_{i,\text{dist}}^{t} + r_{i,\text{col}}^{t} + r_{i,\text{control}}^{t}$.  
$r_{i,\text{dist}}^{t} = - \alpha_{\text{dist}} \norm{p_i^{t}}_2$ encourages robot $i$ to minimize its relative distance to the goal. 
$r_{i,\text{col}}^{t} = -\alpha_{\text{ocol}} \1_{\text{ocol}}^{t} -\alpha_{\text{rcol}} \1_{\text{rcol}}^{t} - \alpha_{\text{rclose}}\sum_{j=1}^{K} \max\left( 1 - \norm{ \Tilde{p}_{ij}^{t}}_2 / d_{\text{rclose}}, 0 \right)\,$ penalizes i) robot-obstacle collisions, ii) inter-robots collisions, and iii) approaching to within $d_{\text{rclose}}$ distance of other robots. 
$\1_{\text{rcol}}^{(t)}$ and $\1_{\text{ocol}}^{(t)}$ are indicator functions, which are equal to $1$ when robots collide with other robots or obstacles, respectively. We only penalize every collision once even though the duration of each collision in the simulation is bigger than one step.
$r_{i,\text{control}}^{t} = - \alpha_{\text{floor}} \1_{\text{floor}}^{t} - \alpha_\omega \norm{\omega_i^{t}}_2 - \alpha_f \norm{f_i^{t}}_2 + \alpha_{\text{orient}} R_{i,33}^{t}\,$ penalizes robot $i$ for i) crashing with the floor, having ii) high angular velocity, iii) high control effort, and iv) big rotation angle relative to the $z$ axis in the world frame. All symbols starting with $\alpha$ and $d$ are hyperparameters.\looseness=-1


\textbf{Reinforcement learning algorithm}:
We use the asynchronous version of the decentralized, independent proximal policy optimization~\cite{schulman2017proximal}. Specifically, we use the  implementation from Sample Factory~\cite{petrenko2020sample} to train our control policies.\looseness=-1

\subsection{Model architecture}
\label{sec:model_architecture}

\begin{figure}
  \centering
    \vspace{0.08in}
    {\includegraphics[width=0.46\textwidth]{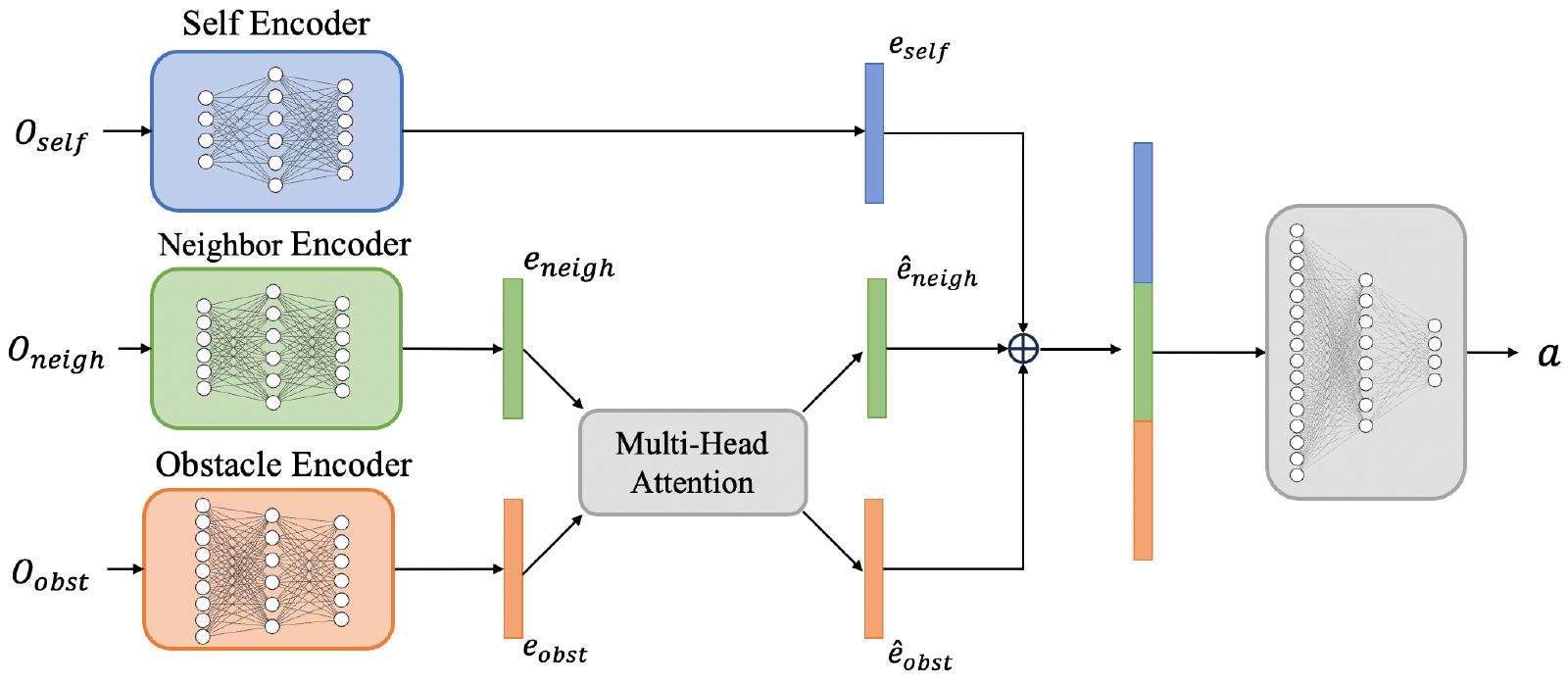}}
    \vspace{-0.1in}
    \caption{\textbf{Model architecture} 
    }
    \label{fig:model_architecture}
    \vspace{-0.28in}
\end{figure}

Our model architecture is a combination of MLPs and a multi-head attention module~\cite{vaswani2017attention}, \mbox{shown in~\autoref{fig:model_architecture}.}
We  use three two-layer MLPs as encoders to process the self, neighbor, and obstacle observations and obtain the corresponding embeddings separately. 
We then fuse these three embeddings.
The multi-head attention module is used to prioritize the importance between neighbor and obstacle embedding: 
$(\Hat{e}_{\text{neigh}}, \Hat{e}_{\text{obst}}) = f_{\text{attn}}(e_{\text{neigh}}, e_{\text{obst}})$. 
Finally, we concatenate $e_{\text{self}}, \Hat{e}_{\text{neigh}}, \Hat{e}_{\text{obst}}$ as the final embedding and feed it into a two-layer MLP to \mbox{obtain the actions $a_i^t$.}\looseness=-1

\subsection{Replay Buffer}
\label{subsec:replay_buffer}
We observe a substantial number of collisions within each episode. 
However, given the time-extended trajectory, these tend to be dispersed and attenuated. 
To amplify collision events and enhance collision aversion toward obstacles and other robots,  once a collision is detected, we append the environment state $\SI{1.5}{s}$ before the collision happens to a replay buffer. 
We save multiple environment states from the same episode if there are multiple collisions during that episode. 
During training, we use a simple curriculum learning method. 
We define the replay rate $\alpha_r$, the probability of replaying one of the previous episodes. 
Alternatively, we generate a new episode with probability $1-\alpha_r$. 
At the end of each episode, we check the number of times each state in the replay buffer is replayed. 
If the replay count of an environment state has exceeded a maximum replay threshold, we assume that the episode starts at this environment state is too difficult to learn for the current policy and we remove it from the replay buffer. This way, all episodes in the buffer are learnable with the current policy.\looseness=-1

\section{Experiments and results}
\label{sec:experiments_and_results}
We evaluate the effectiveness of our learned controller by 
i) ablating its important parts and showing that all parts are required for its effectiveness, 
ii) investigating reward function,
iii) conducting scaling experiments to show its applicability to highly cluttered environments, 
iv) comparing it to two state-of-the-art baselines, and 
v) transferring it to real quadrotors and showing zero-shot sim-to-real transfer. 
vi) investigating generalizability.
The base setting for experiments reported in this section is 8 robots where each robot is able to sense 2 nearest neighbors, $20\%$ obstacle density and $0.6 m$ obstacle size in a $10m\times10m\times10m$ room. 
We train our experiments across 4 different seeds.\looseness=-1

We use five metrics during our evaluations. \textbf{Success rate} is the ratio of robots that reach their goal without collisions, \textbf{collision rate} is the ratio of robots that collide with other robots or obstacles at least once during the whole episode,
\textbf{distance to goal} is the average distance (across all robots) to the goal in the final second of the episode, \textbf{flight distance} is the average flight distance (across all robots) throughout the whole episode, and  \textbf{inference time} is the total inference time from observation collection to emitting actions.\looseness=-1

\subsection{Ablation study}

\begin{figure}
    \centering
    \vspace{0.09in}
    \includegraphics[width=0.48\textwidth]{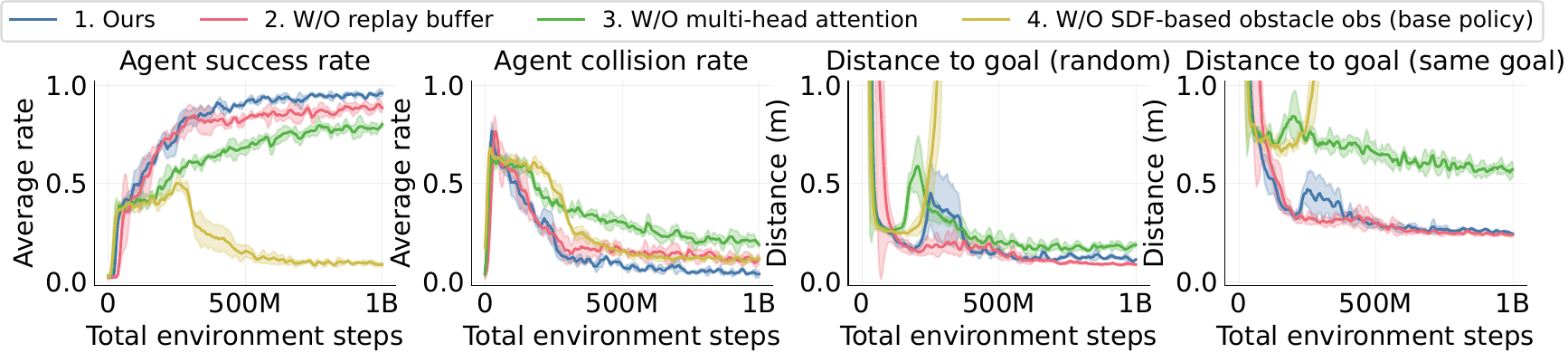}
    \vspace{-0.30in}
    \caption{\textbf{Ablation study:} We remove components one-by-one in order to show the necessity of various parts.
    }
    \label{fig:ablation_study_corl21}
    \vspace{-0.25in}
\end{figure}

We conduct an ablation study to investigate the impact of the new components and present the results in \autoref{fig:ablation_study_corl21}.\looseness=-1 

\begin{figure}
    \vspace{0.09in}
    \centering
    \includegraphics[width=0.48\textwidth]{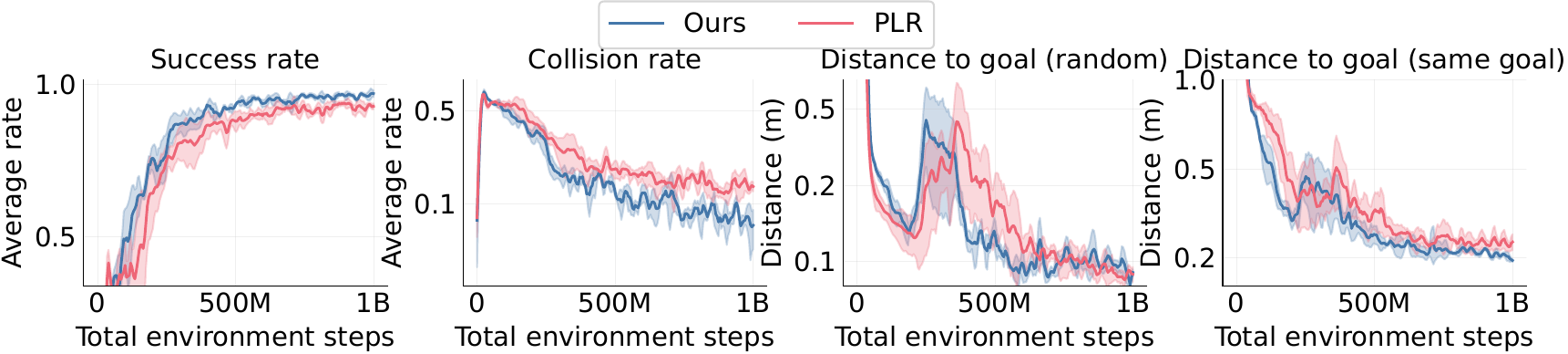}
    \vspace{-0.25in}
    \caption{Comparison of our replay strategy with prioritized level replay.}
    \label{fig:plr_replay}
    \vspace{-0.15in}
\end{figure}

\textbf{Replay mechanism}: 
Removing the replay mechanism results in a noticeable drop in performance. 
For the collision rate plot, we find that without the replay buffer, the collision rate increases from $0.05$ to $0.12$.
Typically when a collision occurs, the collision itself lasts for only one to two timesteps out of a 1500 timestep episode. By storing collision events in a buffer and clipping the episode around the collision event, we essentially force the RL algorithm to focus on learning better actions to reduce the collision penalty by artificially reducing the sparsity of the collision reward.  
To further demonstrate the effectiveness of our replay mechanism, we compare our method with a popular curriculum learning approach, prioritized level replay (PLR)~\cite{jiang2021prioritized}. 
The success rate plot in~\autoref{fig:plr_replay} shows our replay mechanism can learn policies that are $5\%$ higher in success rate than the policies trained with PLR.
Our hypothesis is that utilizing a potential score function (L1 value loss) in PLR encourages faster convergence to the target location \textit{and} collision avoidance, which are not always aligned goals.
In contrast, our replay mechanism solely focuses on collision avoidance.\looseness=-1

\textbf{Multi-head attention}: 
Removing the multi-head attention model results in a further performance drop.
In~\autoref{fig:ablation_study_corl21}, the success rate drops from $0.88$ to $0.79$, the collision rate increases from $0.12$ to $0.20$, distance to goal (random) increases from $0.09m$ to $0.18m$, and distance to goal (same goal) increases from $0.24m$ to $0.57m$.
The attention mechanism enables the agents to prioritize certain agent-agent and agent-obstacle interactions over others, e.g., according to their distances or collision courses.
Without this mechanism, agents equally weigh other objects, which result in low performance.\looseness=-1

\textbf{SDF-based obstacle observations}: 
Switching from the compact and scalable SDF-based obstacle representation to a L-nearest obstacles obstacle representation, we get the policy proposed in~\cite{batra2022decentralized} for inter-robot and dynamic obstacle collision avoidance, which results in even further performance drop.
We find that without the SDF representation, the success rate drops from $0.79$ to $0.10$, representing a \textit{nearly 70$\%$} reduction in performance and a divergence in training.
Distance to goal (random) and distance to goal (same goal) plots convey similar trends. 
L-nearest obstacle representation, although share the same idea as neighbor robots representation, does not work for flying through gaps between obstacles. 
We hypothesize that learning collision avoidance is different from learning to fly through gaps. 
To avoid collisions, robots only need to stay away from neighbor robots or obstacles. However, to fly through gaps, the robots need to precisely estimate their feasible moving spaces. L-nearest obstacle observations, with reduced representational capacity compared to the proposed SDF-based representations, results in a lower ability to navigate obstacle-dense environments, especially when the number of obstacles exceeded the number of nearest-neighbor encoders.\looseness=-1

\begin{figure}
    \centering
    \includegraphics[width=0.48\textwidth]{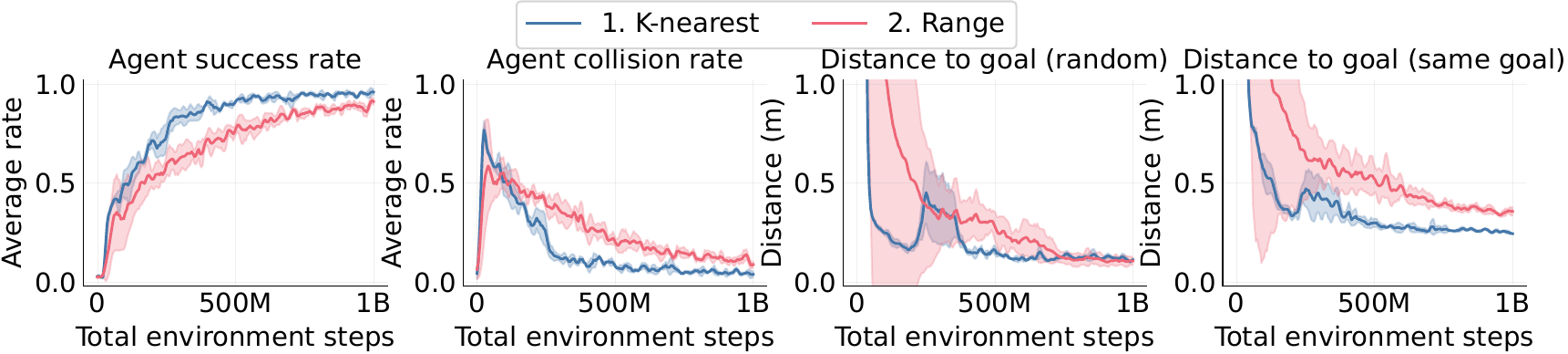}
    \vspace{-0.30in}
    \caption{Comparison of K-nearest neighbor observations with range-based neighbor observations. $K=2$ and range = $4m$.}
    \label{fig:compare_neighbor_obs}
    \vspace{-0.3in}
\end{figure}

\textbf{Range-based neighbor observations}:
We compare our K-nearest neighbor observations with range-based neighbor observations, which sense all neighbor robots in the pre-defined range. 
To support range-based representations, instead of using a two-layer MLP, we use the attention model proposed in~\cite{batra2022decentralized} to deal with varying length of neighbor observations.
In~\autoref{fig:compare_neighbor_obs}, if we replace K-nearest neighbor observations with range-based neighbor observations, the success rate drops from $0.95$ to $0.89$, the collision rate increases from $0.05$ to $0.11$, and the distance to goal (same goal) increases from $0.26m$ to $0.36m$. 
We hypothesize that although range-based observations can provide more information to make robots to perform better in theory, with a more complicated architecture, the learning process becomes more challenging.\looseness=-1

\subsection{Analyzing reward functions}

\begin{figure*}
  \centering
     \subfloat[Before navigating through obstacles]{
         \includegraphics[width=0.30\linewidth]{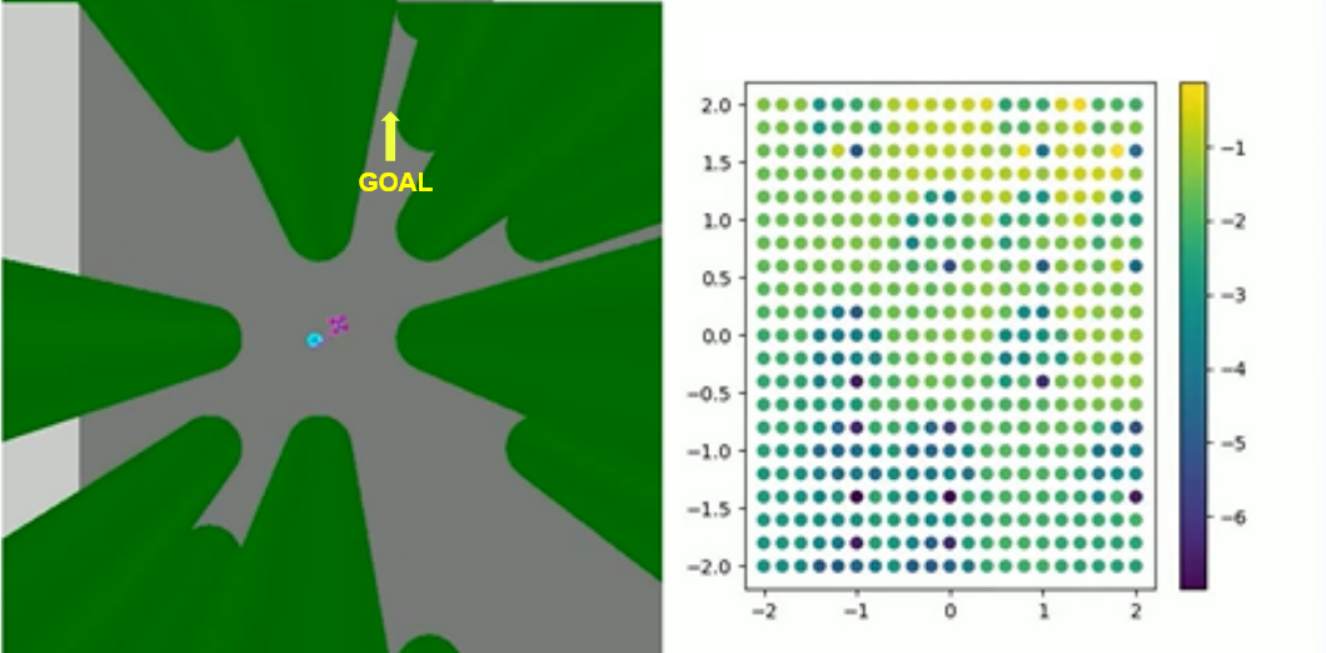}
     }
     \hfill
     \subfloat[Navigating through obstacles]{
        \includegraphics[width=0.30\linewidth]{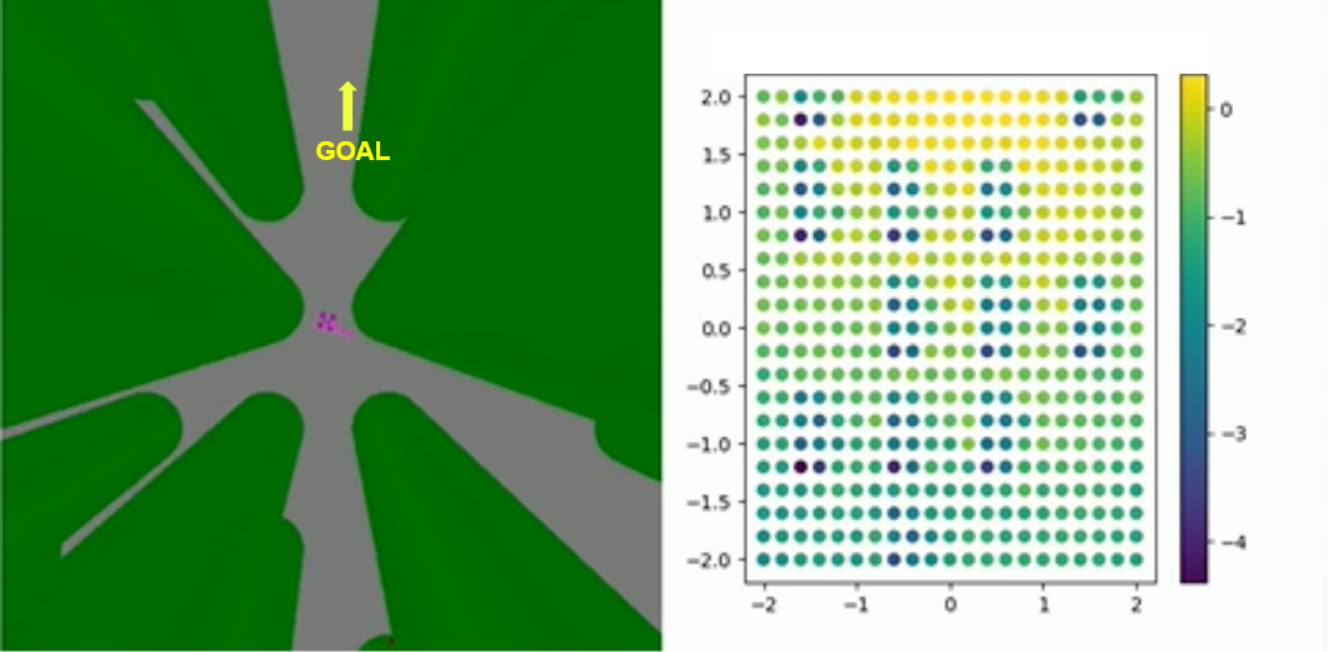}
     }
     \hfill
     \subfloat[After navigating through goal]{
        \includegraphics[width=0.30\linewidth]{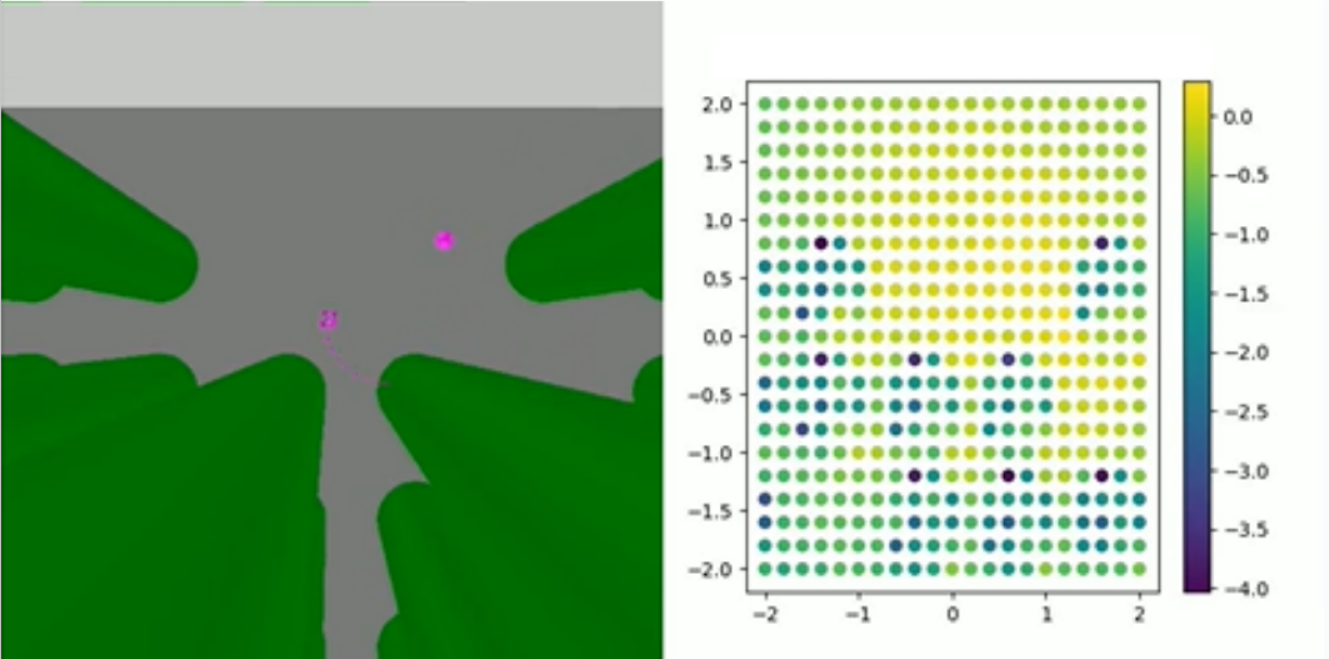}
     }
     \vspace{-0.06in}
    \caption{\textbf{V-Value Maps.} 
    (a), (b), (c) shows the moment before, during, and after the pink quadrotor flying through obstacles. In each subfigure, the left part is a top-down view of the environment, and the right part is a V-value map which is calculated given different positions. 
    The blue sphere in (a) is the goal of other robots. The pink sphere in (c) is the goal of the pink robot.
    }
    \label{fig:v_value}
    \vspace{-0.16in}
\end{figure*}

To analyze if our reward function is designed properly, we use the critic encoder in our training algorithm, IPPO, to evaluate the state value (V-value) by investigating the relation between position and V-value.  
For better visualization, we simplify the position from 3D to 2D and set the velocity of all robots to $0$. 
For each robot, we change its position and change the relative positions to its K-nearest neighbor robots accordingly given the pre-defined resolution. And we use the critic encoder to calculate the V-value at every position in the pre-defined range to build the V-value map. \autoref{fig:v_value} shows the critic encoder is reasonably well in estimating the V-value.\looseness=-1

\subsection{Scaling}
\subsubsection{Number of robots}
\begin{figure}
  \centering
      \vspace{-0.08in}
  \includegraphics[width=0.48\textwidth]{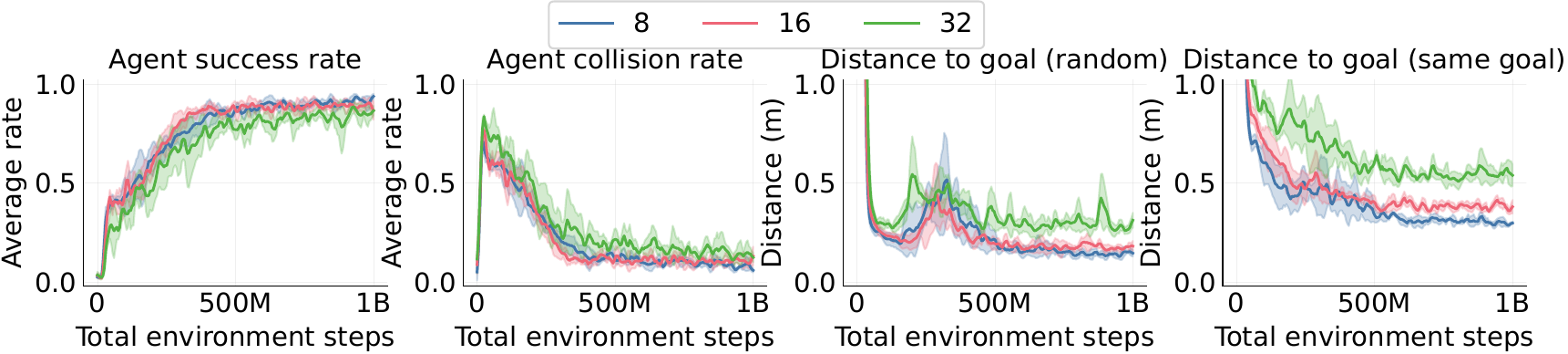}
  \vspace{-0.3in}
  \caption{Experiments with varying number of robots.}
  \label{fig:scale_robots_num}
    \vspace{-0.12in}
\end{figure}

We investigate the scalability of our approach while keeping the sensed number of neighbor robots at 2. The results in \autoref{fig:scale_robots_num} show that our policies can scale up to $32$ robots without a significant decrease in success rate in a $10m\times10m\times10m$ room. With a larger room, we hypothesize that our policies can scale to a greater number of robots.
\looseness=-1

\subsubsection{Number of sensed neighbor robots}

\begin{figure}
  \centering
    \vspace{-0.02in}
  \includegraphics[width=0.48\textwidth]{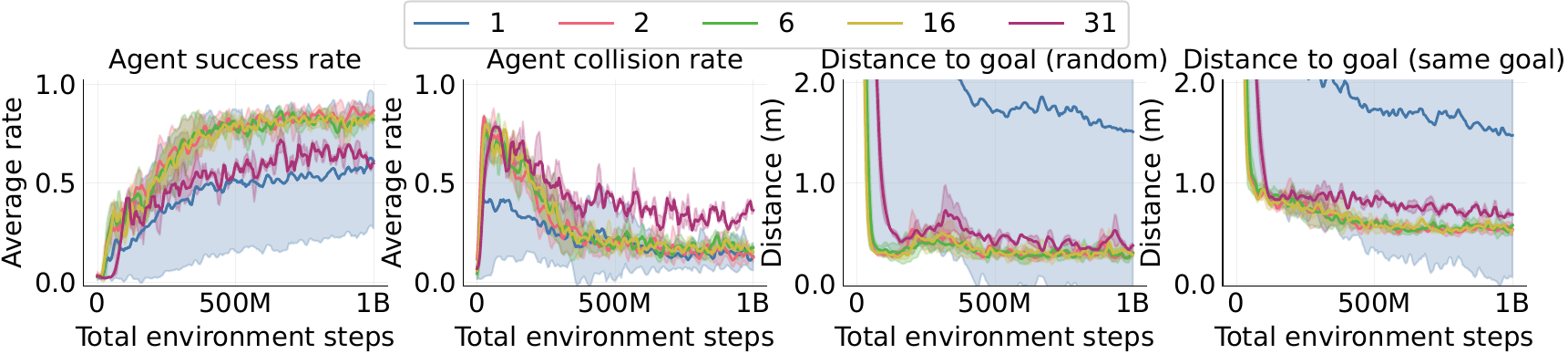}
    \vspace{-0.3in}
  \caption{Experiments with varying sensible number of neighbor robots.}
\label{fig:scale_sensible_neighbor_robots}
    \vspace{-0.28in}
\end{figure}

We fix the number of robots at 32 and set the number of sensed neighbor robots to $1, 2, 6, 16$ and $31$. The results in~\autoref{fig:scale_sensible_neighbor_robots} show our policies work even with one neighbor that can be sensed, but sensing two neighbor robots stably provides the best performance. A larger number of neighbors does not help performance and even decreases performance, such as the $31$ neighbor.
 We attribute this to the significant increase in the input dimension, increasing the hardness of the learning problem.\looseness=-1

\subsubsection{Obstacle density}
\begin{figure}
  \centering
  \vspace{-0.25in}
  \includegraphics[width=0.48\textwidth]{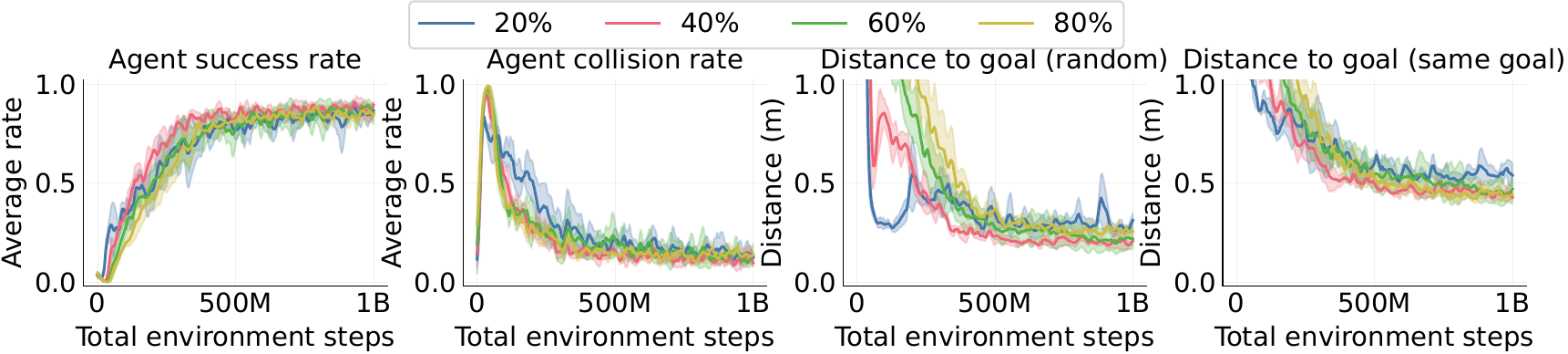}
    \vspace{-0.32in}
  \caption{Experiments with varying obstacle density.}
  \label{fig:scale_obst_density}
    \vspace{-0.08in}
\end{figure}

We investigate the ability of our policies to scale to large obstacle density with the obstacle size fixed at 0.6m. The results in~\autoref{fig:scale_obst_density} show the robustness of our policies to obstacle density scaling; our method can scale up to $80\%$ obstacle density.\looseness=-1

\subsubsection{Obstacle size}
\begin{figure}
  \centering
    \vspace{-0.06in}
  \includegraphics[width=0.48\textwidth]{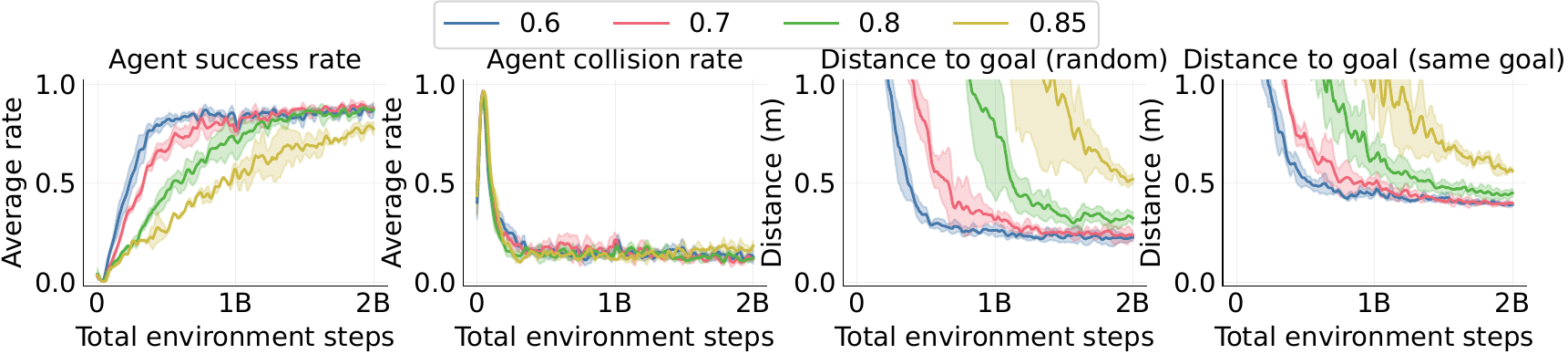}
    \vspace{-0.32in}
  \caption{Experiments with varying obstacle size.}
  \label{fig:scale_obst_size}
\end{figure}

We investigate the ability of our policies to scale to large obstacle size with the obstacle density fixed at $80\%$. The results in~\autoref{fig:scale_obst_size} show our policies can scale up to 0.85m obstacle size.\looseness=-1

\subsection{Baseline comparison}

We compare our approach with a state-of-the-art safety barrier certificates-based method SBC~\cite{wang2017safety} and a state-of-the-art learning-based method GLAS~\cite{riviere2020glas} in simulations.\looseness=-1

SBC is a method using safety barrier certificates, implemented in conjunction with a controller~\cite{mellinger2011snap} that computes safe accelerations to reach the goal position given the current state and then converts the acceleration into thrusts. SBC takes the state of all the neighbors and obstacles within a certain radius of the robot. Given this information and the required acceleration from the controller, SBC outputs a safe acceleration command to direct the robot to the goal, which is then converted intro direct thrust control by the controller. In GLAS, each robot takes the observation of neighbor robots and obstacles within a sensing radius as the input and outputs velocity in $xy$ plane at the next timestep. Since GLAS is implemented in 2D, we set the spawning point and the goal of each robot with the same z-value for a fair comparison. We also use the same controller as above to transform GLAS output into direct thrusts.\looseness=-1 

\begin{table}
\centering
  \vspace{0.09in}
\caption{Baseline Comparison}
  \vspace{-0.15in}
\resizebox{0.48\textwidth}{!}{
\begin{tabular}{lcccc}
\hline
\rule{0pt}{3.5ex}\textbf{Method}  & \makecell{\textbf{Success rate} $\uparrow$} & \makecell{\textbf{Collision rate} $\downarrow$} &  \makecell{\textbf{Flying distance(m)} $\downarrow$}  & \makecell{\textbf{Inference time(ms)} $\downarrow$}\rule[-1.2ex]{0pt}{0pt} \\
\hline
\rule{0pt}{2.5ex}GLAS & 0.75 & 0.25 & 4.4 (2D)  & 15 \\
SBC      & 0.99  &  0.01 & 7.1 (3D) & 21 \\
\textbf{Ours} & 0.97 & 0.03 & 5.3 (3D) & 5
\rule[-1.0ex]{0pt}{0pt} \\
\hline
\end{tabular}
}
\label{table:baseline_comparison}
  \vspace{-0.26in}
\end{table}

In Table~\ref{table:baseline_comparison}, we compare with GLAS and SBC in the environment with $8$ robots and $20\%$ obstacle density. $\uparrow$ represents higher value refers better performance, and $\downarrow$ represents lower value refers better performance. Table~\ref{table:baseline_comparison} shows our method outperforms GLAS by 22\% in terms of success rate and 3 times faster in inference time. Our method is comparable to SBC in success rate and collision rate while being 4x faster in inference speed than SBC.\looseness=-1
\begin{table}
\centering
  \vspace{0.15in}
\caption{Train from scrach vs Policy Distillation}
  \vspace{-0.15in}
\resizebox{0.48\textwidth}{!}{
\begin{tabular}{lccc}
\hline
\rule{0pt}{2.5ex}\textbf{Method}  & \makecell{\textbf{Success rate} $\uparrow$} & \makecell{\textbf{Collision rate} $\downarrow$} &  \makecell{\textbf{Distance to goal(m)} $\downarrow$} \\
\hline
\rule{0pt}{2.5ex}From Scratch        & 0.88  & 0.04  & 0.43 \\
Policy Distillation                  & 0.72           &  0.28            & 0.31\rule[-1.0ex]{0pt}{0pt} \\
\hline
\end{tabular}
}
\label{table:policy_distillation_comparison}
  \vspace{-0.28in}
\end{table}

\begin{figure}
  \centering
    \vspace{-0.08in}
  \includegraphics[width=0.48\textwidth]{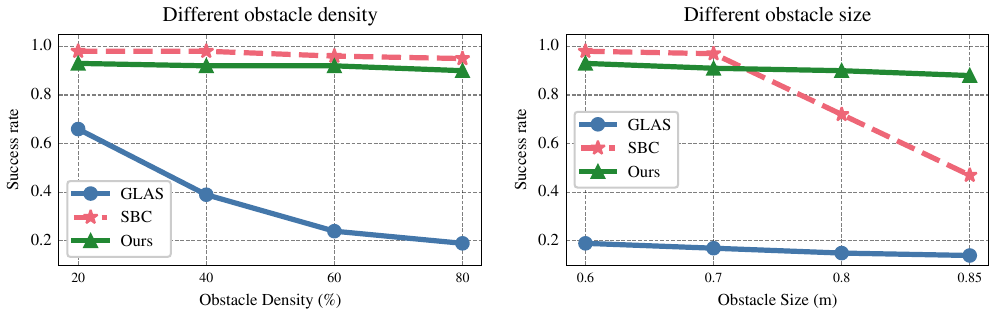}
    \vspace{-0.32in}
  \caption{Baseline comparison: different obstacle densities and obstacle sizes. Obstacle size$=0.6m$ (left). Obstacle density=$80\%$ (right).}
  \label{fig:baseline_comparison}
    \vspace{-0.15in}
\end{figure}

We further compare our approach with these two baselines in more complex environments by fixing the number of robots at 32, and comparing with different obstacle density and obstacle size. 
\autoref{fig:baseline_comparison} shows that our polices are insensitive to obstacle density and obstacle size, while SBC is sensitive to obstacle size, and GLAS is sensitive to obstacle density and does not work when obstacle density is $80\%$ and obstacle size $\geq$ 0.6m. 
When the obstacle size is 0.6m, our policies are comparable to SBC regardless of obstacle density. 
However, in the environment with 32 robots and 80$\%$ obstacle density, when the obactacle size increases to 0.8m, our policies outperform SBC. In this case, if two obstacles are positioned adjacent to each other, there is a gap of only 0.2m between them, and the diameter of drone itself is 0.1m. When we further increase the obstacle size even larger to 0.85m (resulting in the narrowest gap between obstacles measuring only 0.15m), the results in \autoref{fig:baseline_comparison} illustrates that our approach maintains its effectiveness, while SBC falls short. This highlights that our policy is better at flying through complex environments with narrow gaps.\looseness=-1

\subsection{Generalization}

We evaluate our control policies in two unseen scenarios, pursuit evasion and swap goals.
In pursuit evasion scenario, all robots pursuit a same goal, and the trajectory of the goal is based on Bézier curve.
In swap goals scenario, all robots swap their goals after a random period of time.
We evaluate the performance of our control policies over 20 episodes in the environment with 8 robots and 20$\%$ obstacle density. The success rate of pursuit evasion scenario is $0.83$, and the success rate of swap goals scenario is $0.85$.\looseness=-1

\subsection{Physical deployment}

We use Crazyflie 2.1, a quadrotor platform as our testbed for physical deployment. We use Vicon system for localization with the frequency of 100 Hz, and each quadrotor's controller runs at 1000 Hz. For obstacle mapping, we use a list to store the location of static obstacles. 
Quadrotors generate and use local SDFs using the obstacle list online.
Given the compute constraints of Crazyflie 2.1 (168 MHz CPU and 192 Kb RAM), we decrease the model size shown in~\autoref{fig:model_architecture}. 
\begin{figure}
  \centering

   \subfloat[Time lapse of $8$ robots flying to their goals in an area with $5$ obstacles]{
    \includegraphics[width=0.45\linewidth]{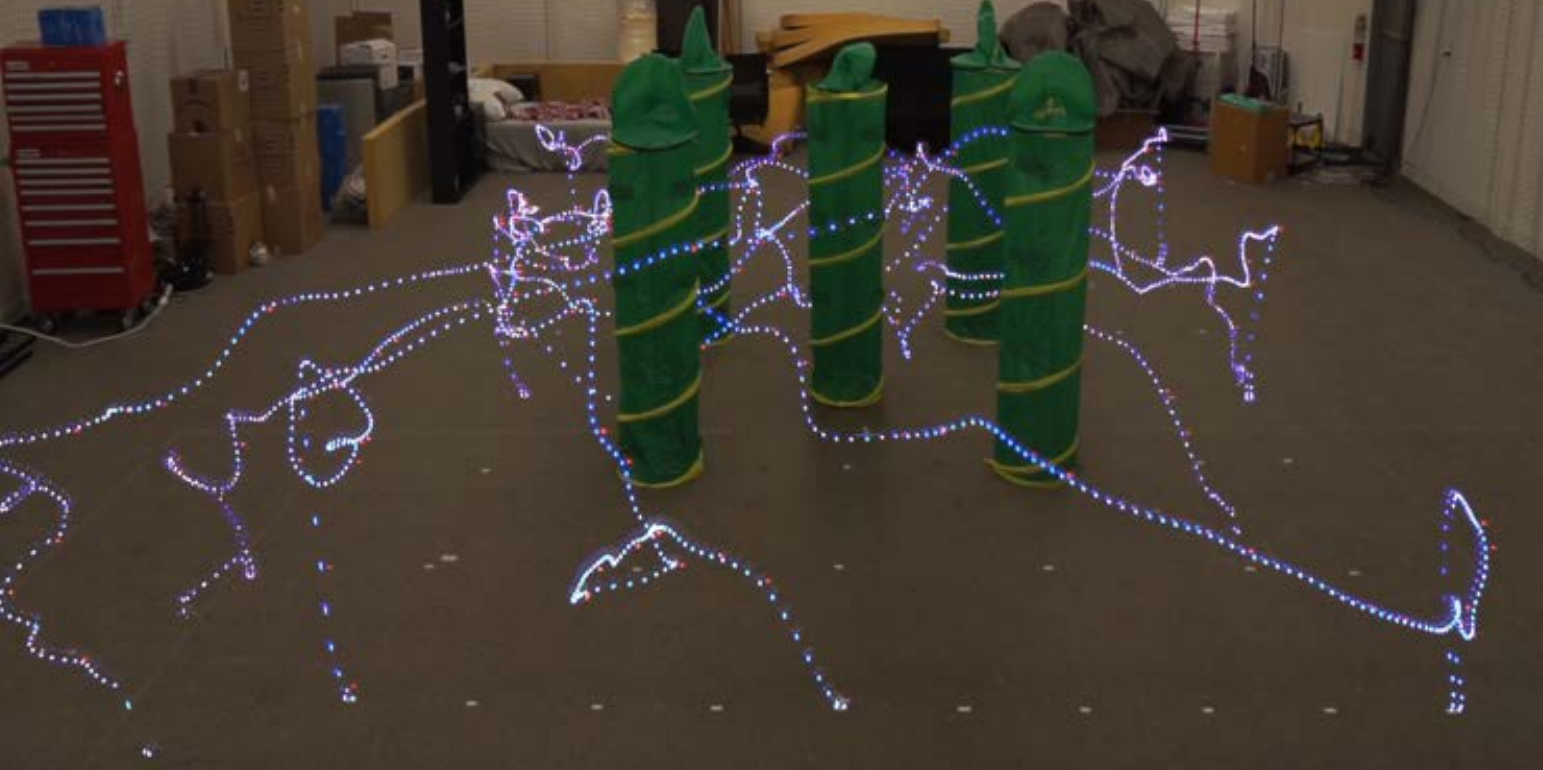}
    \label{fig:time_lapse_8}
   }
   \hfill
   \subfloat[Comparison between simulation and physical deployment.]{
    \includegraphics[width=0.45\linewidth]{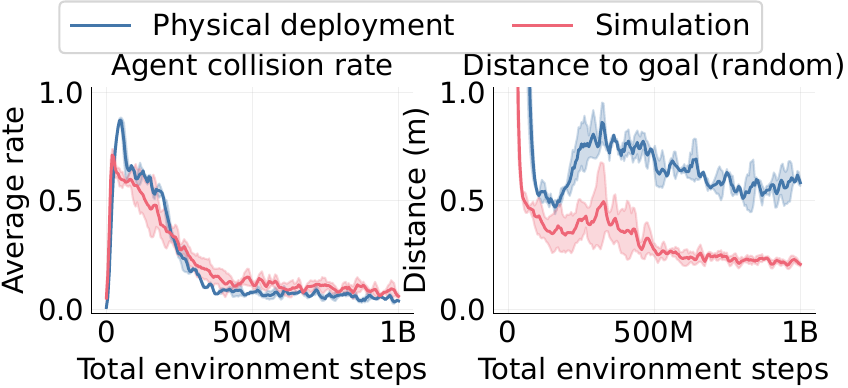}
    \label{fig:one_vs_multi_head_attn}
   }
      \vspace{-0.06in}
  \caption{\textbf{Physical deployment.} The smaller model used in the physical deployment has same collision avoidance performance but with higher distance to goal compared to the simulation model.}
  \label{fig:sim_to_real_compare}
    \vspace{-0.25in}
\end{figure}
We set the hidden size of self encoder, neighbor encoder, obstacle encoder, and attention layers to 10. Further, we replace multi-head attention with single head to reduce the number of operations needed. 
Through these operations, we end up with a model with $1820$ network parameters ($7 KB$ of memory), which runs at $0.35$ ms on-board. 
To train a model that can be deployed on Crazyflie 2.1, we use two methods, training from scratch and using policy distillation~\cite{rusu2015policy}.
We compare their performance with the same model architecture in the environment of 8 robots and $20\%$ obstacle density, and evaluate the performance over 20 episodes. The results are listed in \autoref{table:policy_distillation_comparison}. The model trained from scratch demonstrates superior performance in both success rate and collision rate to the model utilizes policy distillation 
As we prioritize on the collision avoidance capabilities, we use the model trained from scratch for physical deployment. 
\autoref{fig:sim_to_real_compare} shows the model used for the physical deployment has comparable collision avoidance performance to the model used in the simulation, albeit at a cost of a higher average distance to goal.\looseness=-1

\section{Limitations and Future Work}
\label{sec:limitations}

\textbf{More complex environments}: Our method only considers static obstacles. This assumption limits the ability of control robots flying in more unstructured environments. 
In the future, we will add dynamic obstacles into the environment.\looseness=-1 

\textbf{Push towards onboard}: 
Our current localization and obstacle detection is not based on on-board sensors. In the future, we will explore utilizing on-board sensors for localization and obstacle detection. 
Besides, investigating the smallest model can be used for physical deployment~\cite{hegde2023hyperppo} is a interesting direction. 


\textbf{Lack of safety and stability guarantees}: Although our approach shows promising performance in collision avoidance and stability, it is not guaranteed. 
Investigating how to design hybrid methods which combine learning-based methods with classical methods that have safety and stability guarantees is an promising direction.


\section{Conclusion}
\label{sec:conclusion}
In this work, we propose an end-to-end decentralized control policy to control a robot swarm. The policy is trained with RL wherein each robot minimizes the distance to its specified goal while avoiding collisions.
We demonstrate that the learned policy transfers zero-shot to the real world on the highly-constrained Crazyflie 2.1 quadrotor platform.
We make three major improvements to prior work: replay mechanism, multi-head attention, sdf-based representation, resulting in high performance in task completion. 
Our policies scale to 32 robots, in simulation at $80\%$ obstacle density. 
Our method achieves comparable collision avoidance and task completion rates to SBC with 4x faster inference speed and performs considerably better than GLAS.
In future work, we plan to investigate more complex environments, explore long horizon planning problems, and make our policy safety-guaranteed.\looseness=-1

\bibliographystyle{IEEEtran}
\bibliography{root}

\end{document}